\RequirePackage{fix-cm}
\documentclass[twocolumn]{svjour3}          
\smartqed  
\usepackage{graphicx}

\usepackage{epsfig}
\usepackage{graphicx}
\usepackage{amsmath}
\usepackage{amssymb}
\usepackage{algorithm}
\usepackage[noend]{algpseudocode}
\usepackage{varwidth}
\usepackage{float}
\usepackage{lipsum}
\usepackage{booktabs}
\usepackage{multirow}

\begin{document}

\title{Image-Based Geo-Localization Using Satellite Imagery
}


\author{Sixing Hu         \and
        Gim Hee Lee.
}


\institute{S. Hu \at
              Computing 1, 13 Computing Drive, Singapore \\
              \email{hu.sixing@u.nus.edu}
           \and
           G. H. Lee \at
              Computing 1, 13 Computing Drive, Singapore
}

\date{Received: date / Accepted: date}

\maketitle

\begin{abstract}
The problem of localization on a geo-referen-ced satellite map given a query ground view image is useful yet remains challenging due to the drastic change in viewpoint.
To this end, in this paper we work on the extension of our earlier work on the Cross-View Matching Network (CVM-Net) \cite{Hu2018} for the ground-to-aerial image matching task since the traditional image descriptors fail due to the drastic viewpoint change. In particular, we show more extensive experimental results and analyses of the network architecture on our CVM-Net. Furthermore, we propose a Markov localization framework that enforces the temporal consistency between image frames to enhance the geo-localization results in the case where a video stream of ground view images is available. Experimental results show that our proposed Markov localization framework can continuously localize the vehicle within a small error on our Singapore dataset.


\end{abstract}

\section{Introduction}
\label{sec:introduction}

	\begin{figure*}
		\centering
		\includegraphics[width=\linewidth]{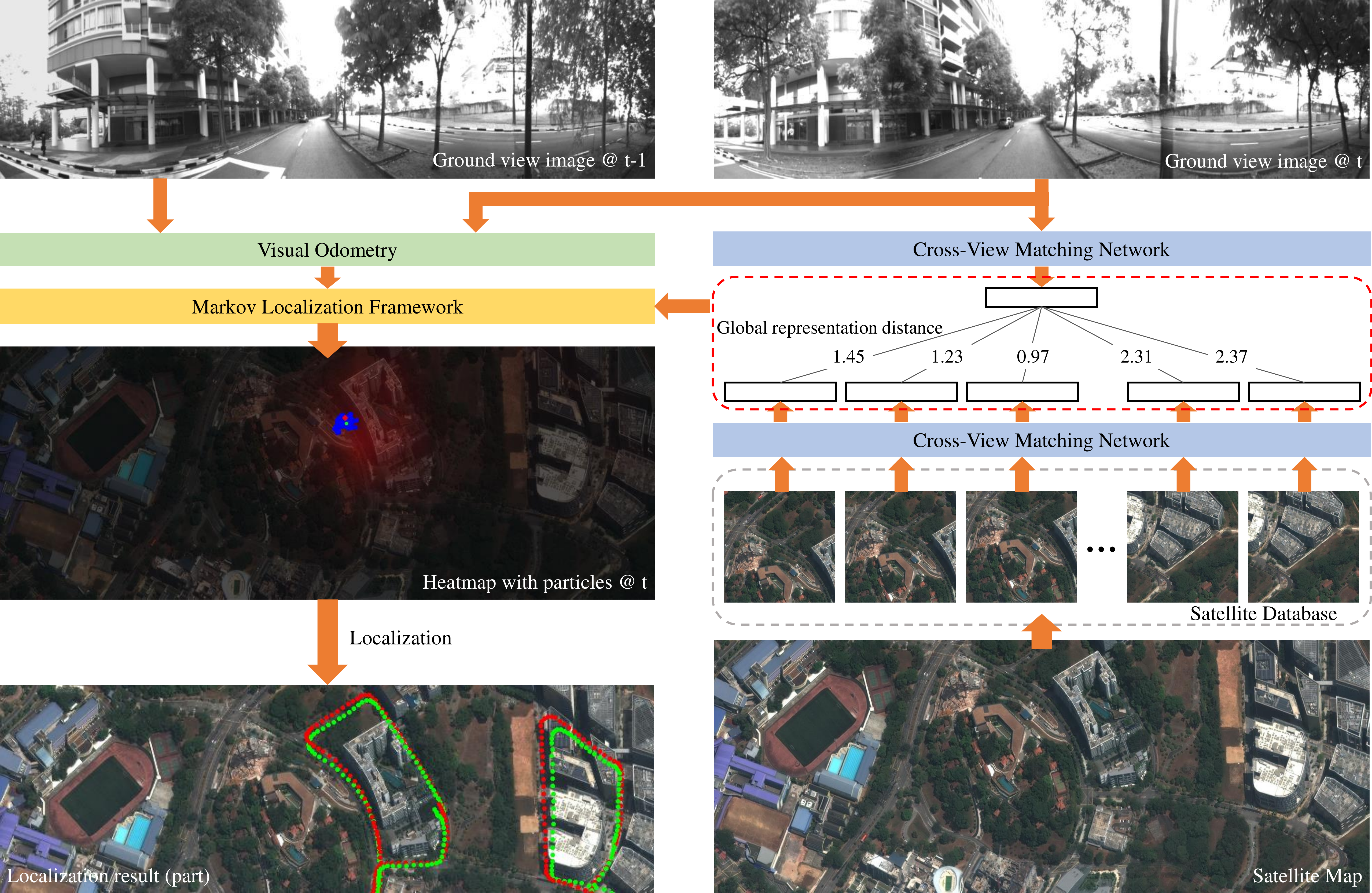}
		\caption {An illustration of the image based ground-to-aerial geo-localization problem, and our proposed framework.}
		\label {fig:teaser}
	\end{figure*}

Image-based geo-localization has drawn a lot of attention over the past years in the computer vision community due to its potential applications in autonomous driving~\cite{Mcmanus2014} and augmented reality~\cite{Middelberg2014}. Traditional image-based geo-localization is normally done in the context where both the query and geo-tagged reference images in the database are taken from the ground view~(\cite{Hays2008}; \cite{Zamir2014}; \cite{Sattler2016}; \cite{Vo2017}). One of the major drawbacks of such approaches is that the database images, which are commonly obtained from crowd-sourcing, e.g. geo-tagged photos from Flickr etc, usually do not have a comprehensive coverage of the area. This is because the photo collections are most likely to be biased towards famous touristy areas. Consequently, ground-to-ground geo-localization approaches tend to fail in locations where reference images are not available. In contrast, aerial imagery taken from devices with bird's eye view, e.g. satellites and drones, densely covers the Earth. As a result, matching ground view photos to aerial imagery gradually becomes an increasingly popular geo-localiza-
\newline
tion approach~(\cite{Bansal2012}; \cite{Lin2013}; \cite{Shan2014}; \cite{Lin2015}; \cite{Workman2015a}; \cite{Workman2015b}; \cite{Vo2016}; \cite{Stumm2016}; \cite{Zhai2017}; \cite{Tian2017}). However, cross-view matching still remains challenging because of the drastic change in viewpoint between ground and aerial images. This causes cross-view matching with traditional handcrafted features like SIFT~\cite{SIFT} and SURF~\cite{SURF} fail.
 
With the recent success of deep learning in many computer vision tasks, most of the existing works on cross-view image matching~(\cite{Workman2015a}; \cite{Workman2015b}; \cite{Vo2016}; \cite{Zhai2017}) adopt the Convolutional Neural Network (CNN) to learn representations for matching between ground and aerial images. To compensate for the large viewpoint difference, Vo and Hays~\cite{Vo2016} use an additional network branch to estimate the orientation and utilize multiple possible orientations of the aerial images to find the best angle for matching across the two views. This approach causes significant overhead in both training and testing. In contrast, our work avoids the overhead by making use of the global VLAD descriptor that was shown to be invariant against large viewpoint and scene changes in the place recognition task~\cite{Jegou2010}. Specifically, we add the NetVLAD layer~\cite{Arandjelovic2016} on top of a CNN to extract descriptors that are invariant against large viewpoint changes. Figure~\ref{fig:teaser} shows an illustration of our approach. The key idea is that NetVLAD aggregates the local features obtained from the CNN to form global representations that are independent of the locations of the local features. We refer to our proposed network as the CVM-Net, i.e. Cross-View Matching Network.

Furthermore, we propose a Markov localization framework, i.e. particle filtering~\cite{Thrun2002}, to achieve global geo-localization of a vehicle running on the road, where a video stream of the ground level images is available. Using the learned representations of ground and satellite images from our CVM-Net, the descriptor distance of a ground view image to all the positions on the satellite map can be computed. The measurement probability for one ground view image of the Markov localization framework is the probability distribution on the satellite map, which is obtained from the descriptor distances. We use the visual odometry computed from consecutive ground images as the basis of the state transition probability distribution in the Markov localization framework. We demonstrate our image-based geo-localization framework on a vehicle equipped with cameras mounted in four orthogonal directions - front, left, rear and right. Experimental results show that our framework is able to localize the vehicle in near real-time with small errors. 

\paragraph{Contributions} This paper is an extension to our earlier work on the CVM-Net~\cite{Hu2018} with two additional contributions: (1) We show extensive experimental results and analyses of our CVM-Net for image-based cross-view ground-to-aerial geo-localization. 
Specifically, we compare the performances of our CVM-Net by replacing the local feature extraction layer with several
recent convolutional architectures~(\cite{Chollet2017Xception}; \cite{Huang2017DenseNet}; \cite{He2016ResNet}). We show experimentally that the VGG architecture~\cite{VGG} is better than other more recent convolutional neural networks on the cross-view image matching task.
(2) Additionally, we propose a Markov localization framework that enforces temporal consistency between image frames from a video stream of ground level images to enhance the geo-localization results of a vehicle moving on the road.
To our best knowledge, our proposed method is the first to achieve near real-time geo-localization of a moving vehicle using only images in a large outdoor area.

\section{Related Work}
\label{sec:related_work}

Most of the existing works on estimating the geographical location of a query ground image used the image matching or image retrieval techniques. These works can be categorized based on the type of features, i.e. hand-crafted and learnable features. There are several existing works that used the Markov localization framework to utilize the temporal information of ground view images to achieve higher localization accuracy.

\paragraph{Hand-crafted features} In the early stage, traditional features that were commonly used in the computer vision community were utilized to do the cross-view image matching~(\cite{Noda2010}; \cite{Bansal2011}; \cite{Senlet2011}; \cite{Senlet2012}; \cite{Lin2013}; \cite{Viswanathan2014}). However, due to the huge difference in viewpoint, the aerial image and ground view image of the same location appeared to be very different. This caused direct matching with traditional local features to fail. Hence, a number of approaches warped the ground image to the top-down view to improve feature matching~(\cite{Noda2010}; \cite{Senlet2011}; \cite{Viswanathan2014}). In cases where building facades are visible
from oblique aerial images, geo-localization can be achieved with facade patch-matching~\cite{Bansal2011}.

\paragraph{Learnable features} As deep learning approaches are proven to be extremely successful in image/video classification and recognition tasks, many efforts were taken to introduce deep learning into the domain of cross-view image matching and retrieval. Workman and Jacobs \cite{Workman2015a} conducted experiments on the AlexNet~\cite{AlexNet} model trained on ImageNet~\cite{Deng2009} and Places~\cite{Zhou2014}. They showed that deep features for common image classification significantly outperformed hand-crafted features. Later on, Workman et al.~\cite{Workman2015b} further improved the matching accuracy by training the convolutional neural network on aerial branch. Vo and Hays~\cite{Vo2016} conducted thorough experiments on existing classification and retrieval networks, including binary classification network, Siamese network and Triplet network. With the novel soft-margin Triplet loss and exhausting mini-batch training strategy, they achieved a significant improvement on the retrieval accuracy. On the other hand, Zhai et al.~\cite{Zhai2017} proposed a weakly supervised training network to obtain the semantic layout of satellite images. These layouts were used as image descriptors to do retrieval from database. 

The most important part of image retrieval is to find a good descriptor of an image which is discriminative and fast for comparison. Sivic and Zisserman~\cite{Sivic2003BoVW} proposed the Bag-of-Visual-Word descriptors to aggregate a set of local features into a histogram of visual words, i.e. the global descriptor. They showed that the descriptor is partially viewpoint and occlusion invariant, and outperformed local feature matching. Nister and Stewenius~\cite{Nister2006} created a tree structure vocabulary to support more visual words. Jegou et al.~\cite{Jegou2010} proposed the VLAD descriptor. Instead of a histogram, they aggregated the residuals of the local features to cluster centroids. Based on that work, Arandjelovic et al.~\cite{Arandjelovic2016} proposed a learnable layer of VLAD, i.e. NetVLAD, that could be embedded into the deep network for end-to-end training. In their extended paper~\cite{Arandjelovic2017}, they illustrated that NetVLAD was better than multiple fully connected layers, max pooling and VLAD. Due to the superior performance of NetVLAD, we adopt the 
\newline
NetVLAD layer in our proposed network. 

\paragraph{Markov Localization}
In many real world applications, e.g. autonomous driving, the ground view images are a stream of video where any image frame is related to its neighboring frames. The Markov localization framework is used in previous works to exploit the temporal relation of the image frames for the cross-view localization task~(\cite{Senlet2011}; \cite{Kim2017}). 
Senlet and Elgammal~\cite{Senlet2011} used the visual odometry results to compute the state transition probability. However, since their measurement probability relies on the matching of the lane marks, it can only be applied on the streets with clear lane marks. Kim and Walter~\cite{Kim2017} used the wheel odometry from the vehicle to compute the state transition probability. They used a deep Siamese network with VGG layers followed by max-pooling to compute the measurement probability. 
 Inspired by these two existing works, we propose a Markov localization framework that can perform vehicle tracking on the geo-referenced satellite map in near real-time and large scale areas with visual odometry as the state transition probability and our CVM-Net as the measurement probability.
 
\paragraph{Image retrieval loss} Our work is also related to metric learning via deep networks. The most widely used loss function in image retrieval task is the max-margin Triplet loss 
that enforces the distances of positive pairs to be less than the distances of negative pairs. The work in~\cite{Hermans2017} concluded that this margin value has to be carefully selected. To overcome this issue, Vo and Hays~\cite{Vo2016} proposed a soft-margin triplet loss which was proven to be effective~\cite{Hermans2017}. Since the triplet loss has no constraint on irrelevant pairs, it will cause the inter-class variation to be small when decreasing the intra-class variation during training. To alleviate this problem, the quadruplet~\cite{Chen2017} and angular~\cite{Wang2017} losses were proposed to further improve the training of triplet network and the performance of image retrieval.


\section{Overview}
\label{sec:overview}
In this paper, we propose a Markov Localization framework, i.e. particle filtering, using a video stream of ground view images to determine the current location of the moving vehicle in a geo-referenced satellite map. The most challenging part in the framework is to compute the measurement probability from the ground view query image and the satellite imagery. In Section~\ref{sec:cvmnet}, we first introduce our cross-view matching network (CVM-Net) and a novel training loss~\cite{Hu2018}. Our CVM-Net extracts the global descriptors of ground view and satellite images. The descriptor distance indicates the similarity of the ground and satellite images. The measurement probability is computed based on the descriptors extracted from our proposed CVM-Net. In Section~\ref{sec:framework}, we introduce the Markov Localization framework for localizing the vehicle moving on the road. The experiments and results are shown in Section~\ref{sec:experiments}, which demonstrate that our proposed Markov Localization framework can localize the vehicle and our proposed deep network is the state-of-the-art architecture for ground-to-aerial cross-view matching.

	\begin{figure*}[t]
		\centering
		\includegraphics[width=\linewidth]{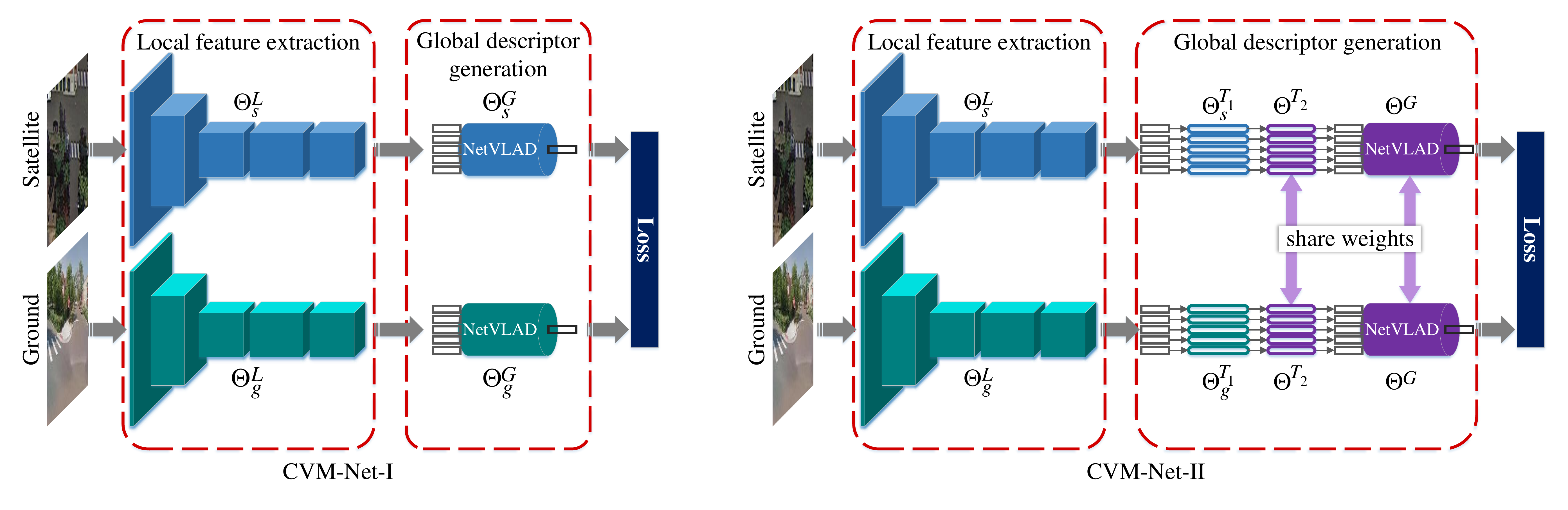}
		\caption {Overview of our proposed CVM-Nets. \textbf{CVM-Net-I}: The deep network with two aligned (no weight-shared) NetVLADs which are used to pool the local features from different views into a common space. \textbf{CVM-Net-II}: The deep network with two weight-shared NetVLADs that transform the local features into a common space before aggregating to obtain the global descriptors.}
		\label {fig:framework}
	\end{figure*}

\section{Cross-View Matching Network}
\label{sec:cvmnet}
Similar to the existing works on image-based ground-to-aerial geo-localization~(\cite{Workman2015b}; \cite{Vo2016}; \cite{Zhai2017}), our goal of the proposed network is to find the closest match of a query ground image from a given database of geo-tagged satellite images, i.e. cross-view image retrieval. To this end, we propose the CVM-Net~\cite{Hu2018}. This section is an extension of our publication~\cite{Hu2018}.

\subsection{Network Overview}
To learn the joint relationship between satellite and ground images, we adopt the Siamese-like architecture that has been shown to be very successful in image matching and retrieval tasks. In particular, our framework contains two network branches of the same architecture. Each branch consists of two parts: local feature extraction and global descriptor generation. In the first part, CNNs are used to extract the local features. See Section~\ref{subsec:local_extraction} for the details. In the second part, we encode the local features into a global descriptor that is invariant across large viewpoint changes. Towards this goal, we adopt the VLAD descriptor by embedding NetVLAD layers on top of each CNN branch. See Section~\ref{subsec:global_generation} for the details.

\subsection{Local Feature Extraction}
\label{subsec:local_extraction}
We use a fully convolutional network (FCN) $f^L$ to extract local feature vectors of an image. For a satellite image $I_s$, the set of local features is given by $U_s = f^L(I_s;\Theta_s^L)$, where $\Theta_s^L$ is the parameters of the FCN of the satellite branch. For a ground image $I_g$, the set of local features $U_g = f^L(I_g;\Theta_g^L)$, where $\Theta_g^L$ is the parameters of the FCN of the ground view branch. In this work, we compare the results of our network using the convolutional part of AlexNet~\cite{AlexNet}, VGG~\cite{VGG}, ResNet~\cite{He2016ResNet}, DenseNet~\cite{Huang2017DenseNet} and Xception~\cite{Chollet2017Xception} as $f^L$. Details of the implementation and comparison are shown in Section~\ref{sec:experiments}.

\subsection{Global Descriptor Generation}
\label{subsec:global_generation}
We feed the set of local feature vectors obtained from the FCN into a NetVLAD layer to get the global descriptor. NetVLAD~\cite{Arandjelovic2016} is a trainable deep network version of VLAD~\cite{Jegou2010}, which aggregates the residuals of the local feature vectors to their respective cluster centroid to generate a global descriptor. The centroids and distance metrics are trainable parameters in NetVLAD. In this paper, we try two strategies, i.e. CVM-Net-I and CVM-Net-II, to aggregate local feature vectors from the satellite and ground images into their respective global descriptors that are in a common space for similarity comparison. 

\paragraph{CVM-Net-I: Two independent NetVLADs}
As shown in Figure~\ref{fig:framework}, we use a separate NetVLAD layer for each branch to generate the respective global descriptors of a satellite and ground image. The global descriptor of an image can be formulated as $v_i = f^G(U_i;\Theta_i^G)$, where $i \in \{s, g\}$ represents the satellite or ground branch. There are two groups of parameters in $\Theta_i^G$ - (1) $K$ cluster centroids $C_i = \{c_{i,1}, ..., c_{i,K} \}$, and (2) a distance metric $W_{i,k}$ for each cluster. The number of clusters in both NetVLADs are set to be same. Each NetVLAD layer produces a VLAD vector, i.e. global descriptor, for the respective views $v_s$ and $v_g$ that are in the same space, which can then be used for direct similarity comparison. More details are given in the next paragraph. To keep computational complexity low, we reduce the dimension of the VLAD vectors before feeding them into the loss function for end-to-end training, or using them for similarity comparison.

\begin{figure}[t]
		\centering
		\includegraphics[width=\linewidth]{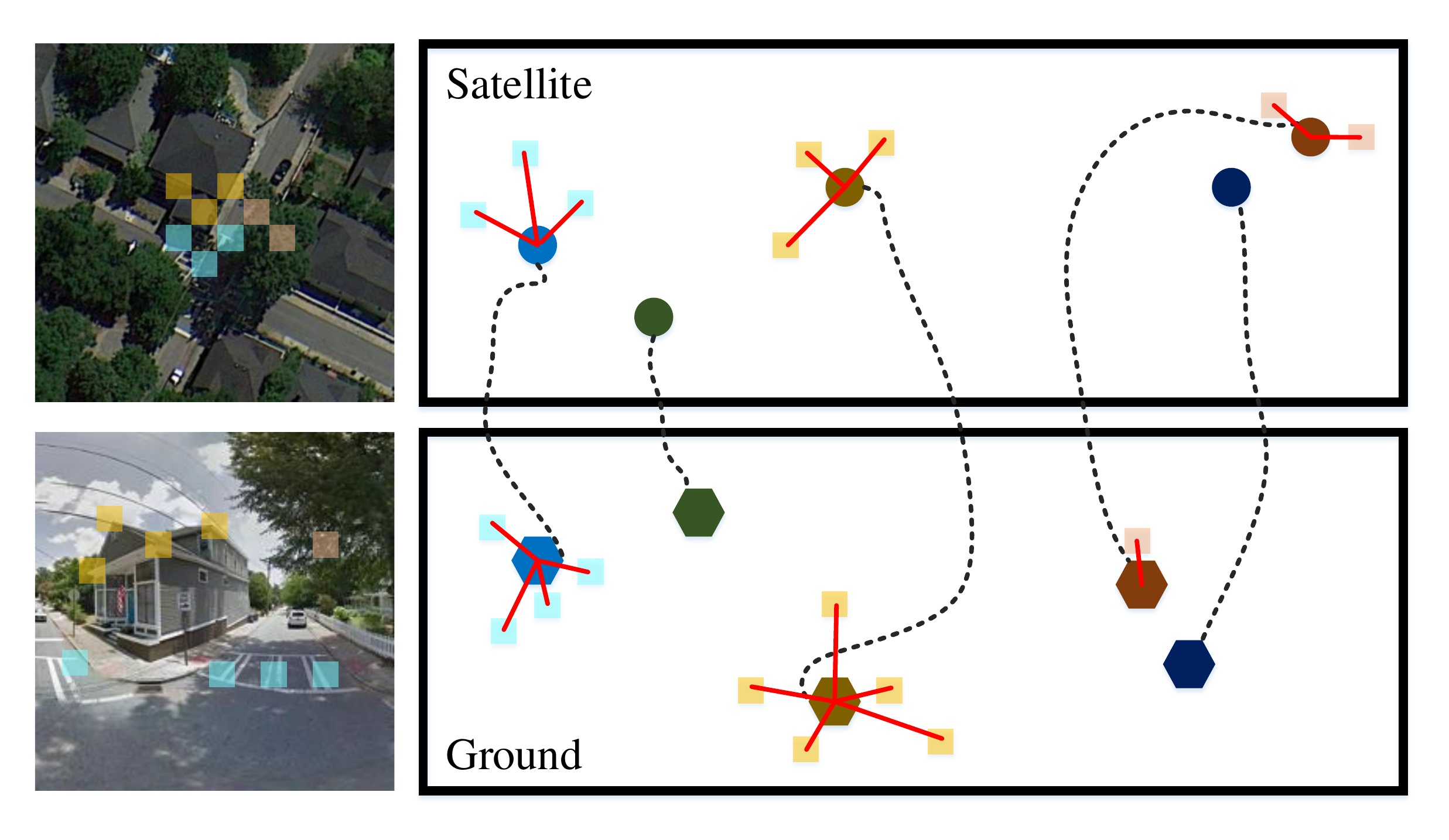}
		\caption{An illustration of how NetVLAD achieves cross-view matching. (Top): satellite view, (Bottom): ground view. In each view, there are a set of local features (colorful squares) and their associated centroids (hexagons and circles). After training, each centroid of satellite view is associated with the unique centroid of ground view (dotted lines). The residuals (red lines) are independent to their own views and comparable to the other view because they are only relative to the centroids. Thus, the global descriptors, i.e. aggregated residuals, of two views are in the common space.}
		\label{fig:vladCrossDomain}
	\end{figure}

In addition to the discriminative power, the two NetVLAD layers with the same number of clusters that are trained together in a Siamese-like architecture, are able to output two VLAD vectors that are in a common space. Given a set of local feature vectors $U = \{u_1, ..., u_N\}$ (we drop the index $i$ in $U_i$ for brevity), the $k^{th}$ element of the VLAD vector $V$ is given by
	\begin{equation}
	\label{eqn:netvlad}
		V(k) = \sum_{j=1}^{N} \bar{a}_k(u_j)(u_j - c_k),
	\end{equation}
where $\bar{a}_k(u_j)$ is the soft-assignment weight determined by the distance metric parameters and input local feature vectors. Refer to \cite{Arandjelovic2016} for more details of $\bar{a}_k(u_j)$. As shown in Equation~\ref{eqn:netvlad}, the descriptor vector of each centroid is the summation of residuals to the centroid. The residuals to the centroids of two views are in a new common space, independent to the domain of two centroids. Therefore, they can be regarded as in a common ``residual'' space with respect to the pair of centroids in two views. The comparison of satellite and ground view descriptors is the centroid-wise comparison. It makes the VLAD descriptors of two views comparable. Figure~\ref{fig:vladCrossDomain} shows an illustration of this concept.

The complete model of our CVM-Net-I is shown in Figure~\ref{fig:framework}. The global descriptor of the satellite image is given by $v_s = f^G(f^L(I_s;\Theta_s^L);\Theta_s^G)$ and ground image is given by $v_g = f^G(f^L(I_g;\Theta_g^L);\Theta_g^G)$. The two branches have identical structures with different parameters. Finally, the dimensions of the global descriptors from the two views are reduced by a fully connected layer.

\paragraph{CVM-Net-II: NetVLADs with shared weights}
Instead of having two independent networks of similar structure in CVM-Net-I, we propose a second network - CVM-Net-II with some shared weights across the Siamese architecture. Figure~\ref{fig:framework} shows the architecture of our CVM-Net-II. Specifically, the CNN layers for extracting local features $U_s$ and $U_g$ remain the same. These local features are then passed through two fully connected layers - the first layer with independent weights $\Theta_s^{T_1}$ and $\Theta_g^{T_1}$, and the second layer with shared weights $\Theta^{T_2}$. The features $U'_s$ and $U'_g$ after the two fully connected layers are given by 
\begin{subequations}
	\begin{align}
	u'_{s,j}&=f^T(u_{s,j}; \Theta_s^{T_1}, \Theta^{T_2}), \label{eqn:fc_sat}\\
	u'_{g,j}&=f^T(u_{g,j}; \Theta_g^{T_1}, \Theta^{T_2}). \label{eqn:fc_grd}
	\end{align}
\end{subequations}
where $u_{s,j} \in U_s$, $u_{g,j} \in U_g$ and $u'_{s,j} \in U'_s$, $u'_{g,j} \in U'_g$.

Finally, the transformed local features are fed into the NetVLAD layers with shared weights $\Theta^G$. The global descriptors of the satellite and ground images are given by
\begin{subequations}
	\begin{align}
	v_s&= f^G(U'_s; \Theta^G), \label{eqn:network2netvlad_sat}\\
	v_g&= f^G(U'_g; \Theta^G). \label{eqn:fc_grd}
	\end{align}
\end{subequations}

The complete model of our CVM-Net-II is illustrated in Figure~\ref{fig:framework}. We adopted weight sharing in our CVM-Net-II network because weight sharing has been proven to improve metric learning in many of the Siamese network architectures, e.g. \cite{Chopra2005}, \cite{Schroff2015}, \cite{Han2015}, \cite{Zagoruyko2015} and \cite{Song2016}.

\subsection{Weighted Soft-Margin Ranking Loss}
The triplet loss is often used as the objective function to train deep networks 
for image matching and retrieval tasks. The goal of the triplet loss is to learn a network that brings positive examples closer to a chosen anchor point than the negative examples. The simplest triplet loss is the max-margin triplet loss: $\mathcal{L}_{max} = max(0, m + d_{pos} - d_{neg})$, where $d_{pos}$ and $d_{neg}$ are the distances of all the positive and negative examples to the chosen anchor. $m$ is the margin and it has been shown in~\cite{Hermans2017} that $m$ has to be carefully selected for best results.
A soft-margin triplet loss was proposed in
to avoid the need to determine the margin in the triplet loss: $\mathcal{L}_{soft} = ln(1 + e^d$), where $d = d_{pos} - d_{neg}$. 
We use the soft-margin triplet loss to train our CVM-Nets, but noted that this loss resulted in slow convergence. To improve the convergence rate, we propose a weighted soft-margin ranking loss which scales $d$ in $\mathcal{L}_{soft}$ by a coefficient $\alpha$:
\begin{equation}
	\label{eqn:our_loss}
	\mathcal{L}_{weighted} = ln(1 + e^{\alpha d}).
\end{equation}

\noindent Our weighted soft-margin ranking loss becomes the soft-margin triplet loss when $\alpha = 1$. We made the observation through experiments that the rate of convergence and results improve as we increase $\alpha$. The gradient of the loss increases with $\alpha$, which might cause the network to improve the weights faster so as to reduce the larger errors.

Our proposed loss can also be embedded into other loss functions with the triplet loss component. The quadruplet loss~\cite{Chen2017} is the improved version of the triplet loss which also tries to force the irrelevant negative pairs further away from the positive pairs. The quadruplet loss is given by
	\begin{equation}
	\label{eqn:quadruplet_loss}
		\begin{split}
		\mathcal{L}_{quad} = & max(0, m_1 + d_{pos} - d_{neg}) + \\
		                     & max(0, m_2 + d_{pos}  -d_{neg}^* ),
		\end{split}
	\end{equation}
\noindent where $m_1$ and $m_2$ are the margins and $d_{neg}^* $ is distance of another example that is outside of the chosen set of positive, negative and anchor examples. We note that the margins are no longer needed with our weighted soft-margin component.  
Our weighted quadruplet loss is given by
	\begin{equation}
	\label{eqn:our_quadruplet_loss}
		\begin{split}
		\mathcal{L}_{quad,weighted} = & ln(1 + e^{\alpha (d_{pos} - d_{neg})}) + \\
		                              & ln(1 + e^{\alpha (d_{pos}  -d_{neg}^* )}).
		\end{split}
	\end{equation}

\section{Image-Based Cross-View Geo-Localization}
\label{sec:framework}

Our proposed CVM-Net described in the previous section provides an effective way to retrieve satellite images from the database given a query ground view image. In this section, we introduce how to use our CVM-Net for geo-localization. 
Despite the effectiveness of our CVM-Net for cross-view matching, our network ignores the temporal consistency of the ground view images from a video stream in autonomous driving. Hence, we propose the Markov Localization framework to enforce temporal consistency between image frames to improve the performance of the ground-to-satellite cross-view localization. More specifically, in this section, we propose the Markov Localization framework, i.e. particle filtering~\cite{Thrun2001}, that recurses over the prediction and update steps, where temporal consistency is enforced via the fusion of visual odometry and cross-view matching results of our CVM-Net from a video stream.

\subsection{Geo-Localization Framework}

The objective of Markov Localization that make use of the particle filtering algorithm is to find the belief distribution, i.e. the posterior probability
of the current vehicle pose $\textbf{x}_t$ given all the past measurements $\textbf{z}_{1:t}$ and control actions $\textbf{u}_t$:
	\begin{equation}
	\label{eqn:belief}
		bel(\textbf{x}_t) = p(\textbf{x}_t | \textbf{z}_{1:t}, \textbf{u}_t).
	\end{equation}
In the particle filter, the belief distribution $bel(\textbf{x}_t)$ is represented by a finite sample set of particles
denoted by:
	\begin{equation}
	\label{eqn:particles}
		\xi_t = \{\chi_t^{[1]}, \chi_t^{[2]}, \cdots, \chi_t^{[M]}\}, 
	\end{equation}
where $\chi_t^{[m]} = [\textbf{x}_t^{[m]}, w_t^{[m]}]^T$ denotes the $m^{th}$ particle. $\textbf{x}_t^{[m]}$ is a random variable that represents the hypothesized state of the $m^{th}$ particle, and $w_t^{[m]}$ is a non-negative value called the importance factor, which determines the weight of each particle.

\begin{algorithm}
	\caption{Image-based cross-view geo-localization}
	\label{algo:localization}
	\begin{algorithmic}[1]
		\Procedure{Localization($I_t$, $I_{t-1}$, $\xi_{t-1}$)}{}
		\State $p(\textbf{z}_t | \textbf{x}_t^{[m]}, \theta) \gets$ Satellite\_Localization($I_t$)
		\State $p(\textbf{x}_t | \textbf{u}_t, \textbf{x}_{t-1}^{[m]}) \gets$ VO\_Localization($I_t$, $I_{t-1}$)
		\State $\xi_t \gets$ PF($\xi_{t-1}$, $p(\textbf{x}_t | \textbf{u}_t, \textbf{x}_{t-1}^{[m]})$, $p(\textbf{z}_t | \textbf{x}_t^{[m]}, \theta)$)
		\State $pose_t \gets$ \textit{average}($\xi_t$)
		\State \textbf{return} $pose_t$
		\EndProcedure
	\end{algorithmic}
\end{algorithm}

Algorithm~\ref{algo:localization} shows the framework of our proposed image-based cross-view geo-localization. The inputs to the framework of each time-stamp is the current ground view image $I_t$, last ground view image $I_{t-1}$ and the most recent particles $\xi_{t-1}$. The framework first computes the measurement probability (denoted as Satellite\_Localization) and the state transition probability (denoted as VIO\_Localization). The new particles are computed through the particle filter algorithm (denoted as PF). The current pose $pose_t$ is the average result of all new particles.
See Sections~\ref{subsec:sat_localization},~\ref{subsec:vo}, and~\ref{subsec:pf} for the details of Satellite\_Localization, VO\_Localization and PF, respectively.


\subsection{Ground-to-Satellite Geo-Localization}
\label{subsec:sat_localization}

We perform image-based ground-to-satellite geo-locali-
\newline
zation with respect to a geo-referenced satellite map with our cross-view image retrieval CVM-Nets. The satellite map is discretized into a database of image patches centered on every $P$ pixels of the map. A smaller interval $P$ gives better localization accuracy with the trade-off of a higher search complexity due to the larger database. 
To balance the localization accuracy and the computation speed, we choose $P = 10$ in our experiment. All satellite images in the database are fed into the satellite branch of our CVM-Net and the descriptors are stored. To query a ground view image, the descriptor is first computed by the ground branch of the CVM-Net. Next, the Euclidean distance $d_{g\text{-}s_i}$ of the global descriptor to the global descriptor of the every satellite image $I_{s_i}$ is calculated. We define the probability of the query image at a location $\boldsymbol{l}_i$ in the satellite map as 
	\begin{equation}
	\label{eqn:loc_confidence}
		p_{\boldsymbol{l}_i} = \dfrac{e^{-d_{g\text{-}s_i}}} {\sum_i e^{-d_{g\text{-}s_i}}},
	\end{equation}

\noindent where the probability of the query image location is smaller for a larger descriptor distance. 
In the particle filter algorithm, the sensor measurement probability distribution $p(\textbf{z}_t | \textbf{x}_t^{[m]}, \theta)$ is obtained from the location probability $p_{\boldsymbol{l}_i}$. $\theta$ represents the given satellite map that the vehicle is working in. For a hypothetical state $\textbf{x}_{t-1}^{[m]} = [x, y, \theta]^T$, we find the 4 nearest location $\{\boldsymbol{l}_{i_1}, \boldsymbol{l}_{i_2}, \boldsymbol{l}_{i_3}, \boldsymbol{l}_{i_4}\}$ to the state location $[x, y]^T$. The probability $p(\textbf{z}_t | \textbf{x}_t^{[m]}, \theta)$ is obtained by bilinear interpolation from these 4 nearest grid corners:
	\begin{equation}
	\label{eqn:loc_model}
		p(\textbf{z}_t | \textbf{x}_t^{[m]}, \theta) = \sum_{j=1}^{4} p_{\boldsymbol{l}_{i_j}}.
	\end{equation}

\subsection{Visual Odometry}
\label{subsec:vo}

We use the visual odometry technique to estimate the relative camera motion from two consecutive key frame images. 
The relative pose computed from the visual odometry is very accurate
for a relatively short range. In this work, we use the visual odometry algorithm proposed by Liu et al.~\cite{Liu2018}. It is the latest and state-of-the-art visual odometry algorithm. In contrast to previous methods, Liu et al. use a multi-camera system to improve the robustness of visual odometry. 
There are two parts in their proposed visual odometry pipeline: tracker and local mapper. The tracker estimates the vehicle pose using the motion predictor and the direct image alignment to the latest key frame. The local mapper estimates the 3D point cloud from the stereo cameras and refines the camera pose and the 3D point cloud to minimize the long-term pose drift. 

In the particle filter algorithm, the random variable $\textbf{x}_t^{[m]}$ is sampled from the motion model $p(\textbf{x}_t | \textbf{u}_t, \textbf{x}_{t-1}^{[m]})$. We compute the state transition probability distribution
$p(\textbf{x}_t | \textbf{u}_t, \textbf{x}_{t-1}^{[m]})$ based on the result of visual odometry. The inputs are the current control data $\textbf{u}_t$ and a hypothetical state $\textbf{x}_{t-1}^{[m]}$ of the vehicle at $t-1$. The control actions $\textbf{u}_t$ are the relative motion information provided by the visual odometry readings of the vehicle and is given by $\textbf{u}_t = [\delta_{trans}, \delta_{rot}]$, where $\delta_{trans}$ is the translated distance and $\delta_{rot}$ is the rotated angle when the vehicle advances from pose $\textbf{x}_{t-1}^{[m]}$ to $\textbf{x}_{t}^{[m]}$ in the time interval $(t - 1, t]$.

The control actions $\textbf{u}_t$ provided by the visual odometry readings are corrupted by noise, which we assume to be Gaussian noise. The ``true'' value of the translation $\hat{\delta}_{trans}$ and rotation $\hat{\delta}_{rot}$ are obtained from $\delta_{trans}$ and $\delta_{rot}$ by subtracting Gaussian noise with zero mean and standard deviation denoted by $\sigma_{trans}$ and $\sigma_{rot}$ for translation and rotation respectively. The current pose $\textbf{x}_{t}^{[m]} = [x, y, \theta]^T$ of the vehicle is computed from its previous pose $\textbf{x}_{t-1}^{[m]}$ and the ``true'' translation $\hat{\delta}_{trans}$ and ``true'' rotation $\hat{\theta}_{rot}$:
	\begin{align}
	\label{eqn:vo_model}
		x &= x_{t-1}^{[m]} + \hat{\delta}_{trans} \ cos(\theta_{t-1}^{[m]} + \hat{\theta}_{rot}), \\
		y &= y_{t-1}^{[m]} + \hat{\delta}_{trans} \ sin(\theta_{t-1}^{[m]} + \hat{\theta}_{rot}), \\
		\theta &= \theta_{t-1}^{[m]} + \hat{\theta}_{rot}.
	\end{align}

\subsection{Particle Filter}
\label{subsec:pf}

\begin{algorithm}
	\caption{Particle filter}
	\label{algo:pf}
	\begin{algorithmic}[1]
		\Procedure{PF($\xi_{t-1}$, $p(\textbf{x}_t | \textbf{u}_t, \textbf{x}_{t-1}^{[m]})$, $p(\textbf{z}_t | \textbf{x}_t^{[m]}, \theta)$)}{}
		\State $\overline{\xi}_t \gets \emptyset$
		\State $\xi_t \gets \emptyset$
		\For {$m=1$ to $M$}
		\State \textit{sample} $\textbf{x}_t^{[m]} \sim p(\textbf{x}_t | \textbf{u}_t, \textbf{x}_{t-1}^{[m]})$
		\State $w_t^{[m]} \gets p(\textbf{z}_t | \textbf{x}_t^{[m]}, \theta)$
		\State $\overline{\chi}_t^{[m]} \gets [\textbf{x}_t^{[m]}$ $w_t^{[m]}]^T$
		\EndFor
		\State $\xi_t \gets$ \textit{resample}($\overline{\xi}_t$)
		\State \textbf{return} $\xi_t$
		\EndProcedure
	\end{algorithmic}
\end{algorithm}

Algorithm~\ref{algo:pf} shows the pseudo code for the particle filter algorithm. The inputs to the algorithm are the previous particle set $\xi_{t-1}$, the state transition probability distribution $p(\textbf{x}_t | \textbf{u}_t, \textbf{x}_{t-1}^{[m]})$ and the sensor measurement probability distribution $p(\textbf{z}_t | \textbf{x}_t^{[m]}, \theta)$. The particle filter algorithm first generates a temporary particle set $\overline{\xi}_t$ that represents the predicted belief distribution $\overline{bel}(\textbf{x}_t)$ in the prediction step. It is then followed by the update step that transforms the predicted belief distribution $\overline{bel}(\textbf{x}_t)$ into the posterior belief distribution $bel(\textbf{x}_t)$. In detail:

\paragraph{Prediction:} Line 5 of the algorithm generates the hypothetical state $\textbf{x}_t^{[m]}$ by sampling from the state transition probability distribution $p(\textbf{x}_t | \textbf{u}_t, \textbf{x}_{t-1}^{[m]})$. The state transition probability $p(\textbf{x}_t | \textbf{u}_t, \textbf{x}_{t-1}^{[m]})$ is obtained from the visual odometry motion model. The set of particles obtained after $M$ iterations is the discrete representation of the predicted belief $\overline{bel}(\textbf{x}_t)$.

\paragraph{Update:} The update step of the particle filter algorithm consists of two steps: importance factor and resampling. The importance factor $w_t^{[m]}$ for the $m^{th}$ particle at time $t$ is computed in Line 6 of the algorithm. Importance factors are used to incorporate the measurement $\textbf{z}_t$ into the particle set and the importance factor of the $m^{th}$ particle is given by the measurement probability $p(\textbf{z}_t | \textbf{x}_t^{[m]}, \theta)$. It should be noted that the particles with hypothetical states closer to the posterior belief distribution $bel(\textbf{x}_t)$ have a higher importance factor.

The resampling step in Line 8 of the algorithm is an important part of the particle filter algorithm. Resampling draws with replacement $M$ particles from the temporary set $\overline{\xi}_t$. The probability of drawing each particle is given by its importance weight. This means that the particles with higher importance weight will have a higher chance of appearing in $\xi_t$. Consequently, the particles will be approximately distributed according to the posterior belief distribution $bel(\textbf{x}_t) = \eta \ p(\textbf{z}_t | \textbf{x}_t) \ \overline{bel}(\textbf{x}_t)$ after the resampling step. The $\eta$ is a normalization factor.

\section{Experiments and Results}
\label{sec:experiments}

\subsection{Dataset and Platform}
\begin{figure}[t]
		\centering
		\includegraphics[width=\linewidth]{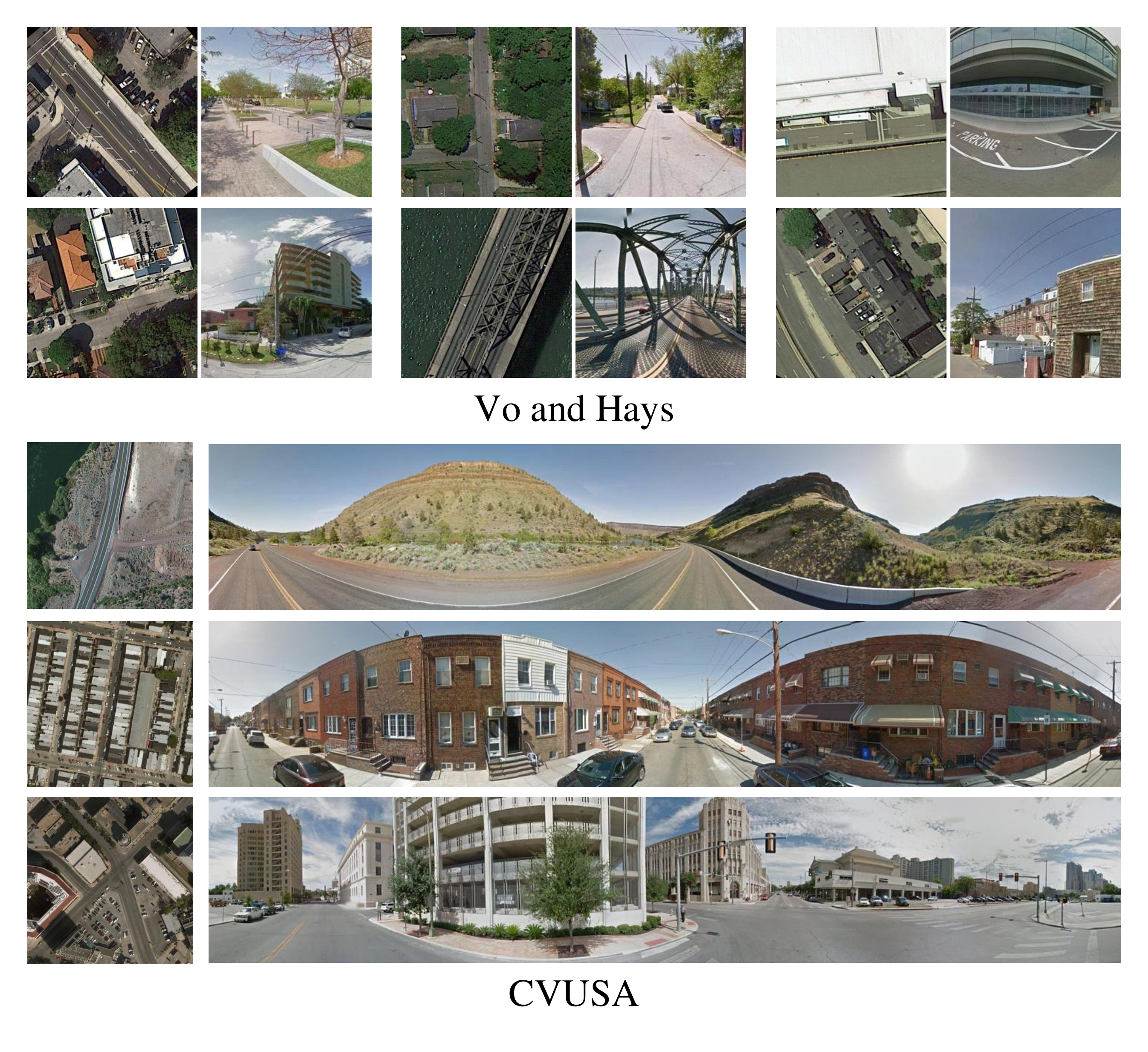}
		\caption {Sample images from the Vo and Hays~\cite{Vo2016}, and CVUSA~\cite{Zhai2017}.}
		\label {fig:dataset}
	\end{figure}

We evaluate our proposed deep networks: CVM-Net-I and CVM-Net-II on two existing datasets -  CVUSA~\cite{Zhai2017} and Vo and Hays~\cite{Vo2016}. The CVUSA dataset contains 35,532 image pairs for training and 8,884 image pairs for testing. All ground images are panoramas. Vo and Hays' dataset consists of around one million image pairs from 9 different cities. All ground images are cropped from panoramic images to a fixed size. We use all image pairs from 8 of the 9 cities to train the networks and use the image pairs from the 9$^{\text{th}}$ city, i.e. Denver city, for evaluation. Figure~\ref{fig:dataset} shows some examples of the two datasets.

\begin{figure}
		\centering
		\includegraphics[width=0.95\linewidth]{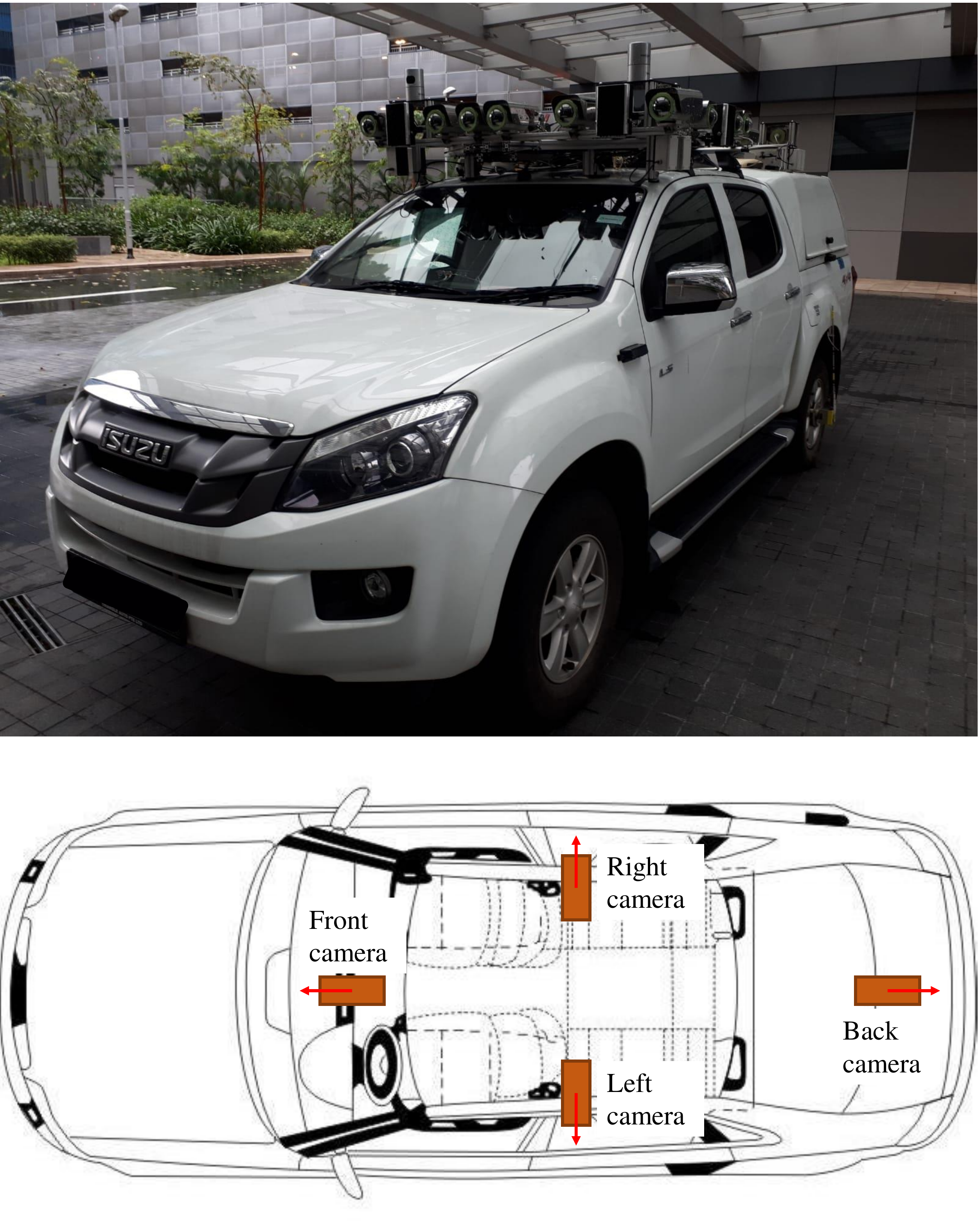}
		\caption {Our experimental vehicle (top) and the cameras used in the experiments (bottom).}
		\label {fig:vehicle}
	\end{figure}

\begin{figure*}[t]
		\centering
		\includegraphics[width=\linewidth]{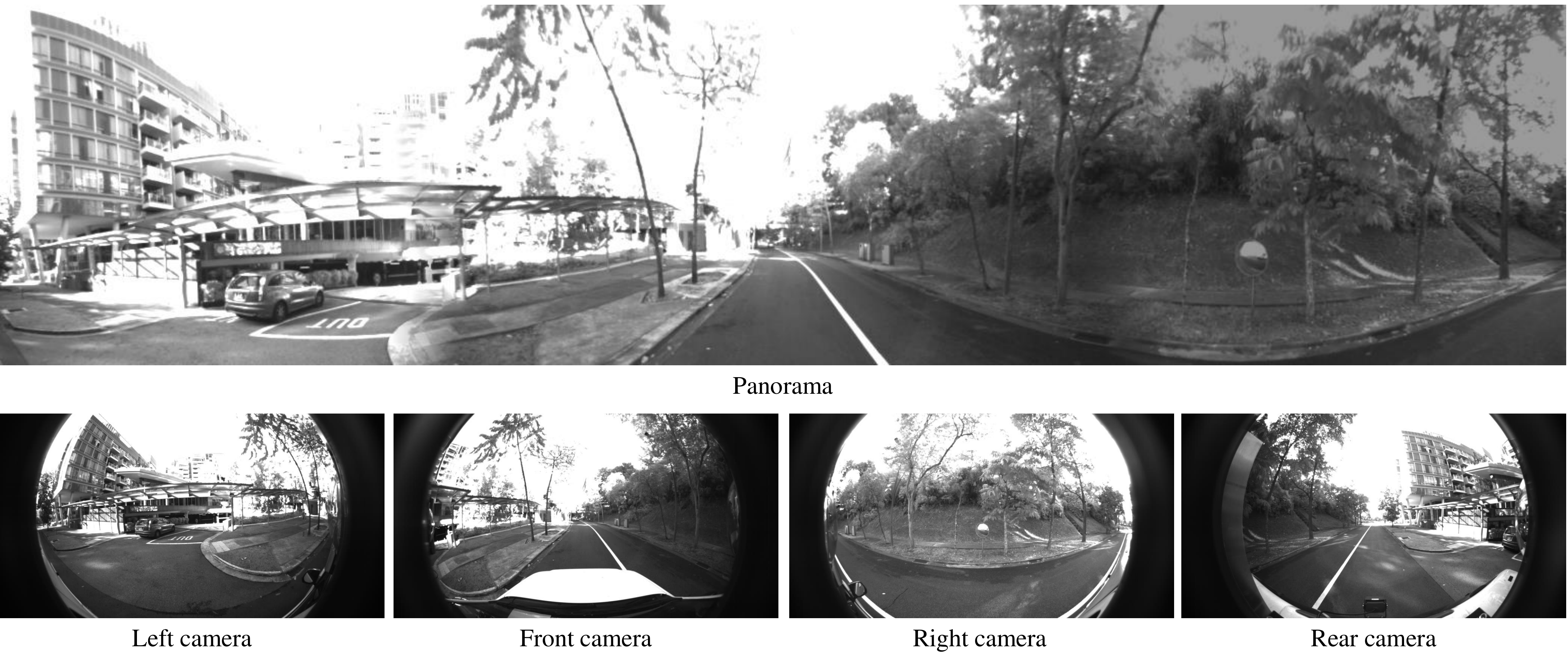}
		\caption {A set of sample images captured from our vehicle. The panorama at the top is obtained by stitching the 4 images at the bottom. 
		}
		\label {fig:sg_dataset}
	\end{figure*}

Our experimental platform for the Markov Localization is a vehicle with 12 fisheye near-infared (NIR) cameras mounted on the top. We use 4 of them which head 4 directions (front, rear, left and right) to form the panoramas. Each camera has a 180-degree field of view. The images from 4 cameras are unwarpped to a cylinder to form a panoramic image. Figure~\ref{fig:vehicle} shows our vehicle and Figure~\ref{fig:sg_dataset} shows an example of images captured from its cameras. It is equipped with GNSS/INS to provide the ground-truth poses.

\subsection{CVM-Net Implementation and Training}
We use the VGG16~\cite{VGG} architecture with 13 convolutional layers to extract local features, and a NetVLAD with 64 clusters to generate the global descriptors. We use VGG16~\cite{VGG} as the local feature extraction network is because it is well-studied and known as one of the best network for local features. The comparisons of other convolutional architectures are provided later. We set $\alpha = 10$ for both the weighted triplet and weighted quadruplet losses. We use the squared Euclidean distance in our loss functions. The parameters in VGG16 are initialized with a pre-trained model on ImageNet~\cite{Deng2009}. All the parameters in NetVLAD and fully connected layers are randomly initialized.

We implement our CVM-Nets using Tensorflow~\cite{Tensorflow} and train using the Adam optimizer~\cite{Kingma2014} with the learning rate of $10^{-5}$ and dropout ($=0.9$) for all fully connected layers. The training is divided into two stages. In the first stage, we adopt the exhaustive mini-batch strategy~\cite{Vo2016} to maximize the number of triplets within a batch. We feed pairs of corresponding satellite and ground images into our Siamese-like architecture. We have a total of $M \times 2(M-1)$ triplets for $M$ positive pairs of ground-to-satellite images. This is because for each ground or satellite image in $M$ positive pairs, there are $M-1$ corresponding negative pairs from all the other images, i.e. $2(M-1)$ for both the ground and satellite images in a positive pair. Once the loss stops decreasing, we start the second stage with in-batch hard negative mining. For each positive pair, we choose the negative pair with smallest distance in current batch.

\subsection{Results of Image Retrieval}

\begin{table}
	\centering
	\caption{Comparison of top 1\% recall on our CVM-Nets with other existing approaches
	and two baselines, i.e. Siamese network with AlexNet and VGG.}
	\begin{tabular}{l|cc}
		\hline
		\multirow{2}{*}{}                  & \multicolumn{2}{c}{Recall @top 1\%} \\
		\cline{2-3}
		                                   & Cropped~\cite{Vo2016}  & Panorama~\cite{Zhai2017}  \\
		\hline \hline
		Siamese (AlexNet)                  & 1.1\%           & 4.7\%             \\
		\hline
		Siamese (VGG)                      & 1.3\%           & 9.9\%             \\
		\hline
		Workman et al. \cite{Workman2015b} & 15.4\%          & 34.3\%            \\
		\hline
		Vo and Hays \cite{Vo2016}          & 59.9\%          & 63.7\%            \\
		\hline
		Zhai et al. \cite{Zhai2017}        & \textemdash     & 43.2\%            \\
		\hline
		CVM-Net-I                          & \textbf{67.9\%} & \textbf{96.3\%}   \\
		\hline
		CVM-Net-II                         & 66.6\%          &  87.2\%           \\
		\hline
	\end{tabular}
	\label{tab:performance_compare}
\end{table}

\paragraph{Evaluation metrics}
We follow Vo and Hays~\cite{Vo2016}, and Workman et al.~\cite{Workman2015b} in using the recall accuracy at top 1\% as the evaluation metric for our networks. For a query ground view image, we retrieve the top 1\% closest satellite images with respect to the global descriptor distance. It is regarded as correct if the corresponding satellite image is inside the retrieved set.

\paragraph{Comparison to existing approaches}
We compare our proposed CVM-Nets to three existing works~\cite{Vo2016,Workman2015b,Zhai2017} 
on the two datasets provided by \cite{Vo2016} and \cite{Zhai2017}. We used the implementations given in the authors' webpages. Furthermore, we take the Siamese network with both AlexNet~\cite{AlexNet} and VGG~\cite{VGG} as the baseline in our comparisons, since these networks are widely used in image retrieval tasks. The AlexNet is used in \cite{Vo2016}.
We use our weighted soft-margin ranking loss in our CVM-Nets. The soft-margin triplet loss is used on the network from Vo and Hays~\cite{Vo2016}, as suggested by the authors in the paper. We also apply the soft-margin triplet loss on the two baseline Siamese networks - AlexNet and VGG since the soft-margin triplet loss produces the state-of-the-art results in~\cite{Vo2016}. The Euclidean loss is used on the network proposed by Workman et al.~\cite{Workman2015b} since their network is trained on only positive pairs. 

Table~\ref{tab:performance_compare} shows the top 1\% recall accuracy results of our CVM-Nets compared to all the other approaches on the two datasets - which we called ``Cropped" \cite{Vo2016} and ``Panorama" \cite{Zhai2017} in the table for brevity. It can be seen that both our proposed networks - CVM-Net-I and CVM-Net-II significantly outperform all the other approaches. This suggests that NetVLAD used in both our CVM-Nets is capable of learning much more discriminative features compared to the CNN and/or fully connected layers architectures utilized by the other approaches. Furthermore, it can be seen that CVM-Net-I outperforms CVM-Net-II in both datasets. 
This result suggests that although weight sharing based Siamese networks performed well in traditional image retrieval tasks, e.g. face identification, it is not necessarily good for our network on cross-view image retrieval. 
It is also not surprising that all networks perform better on the panorama images since these images contain more information from the wide field-of-views. 

\begin{figure}
		\centering
		\includegraphics[width=\linewidth]{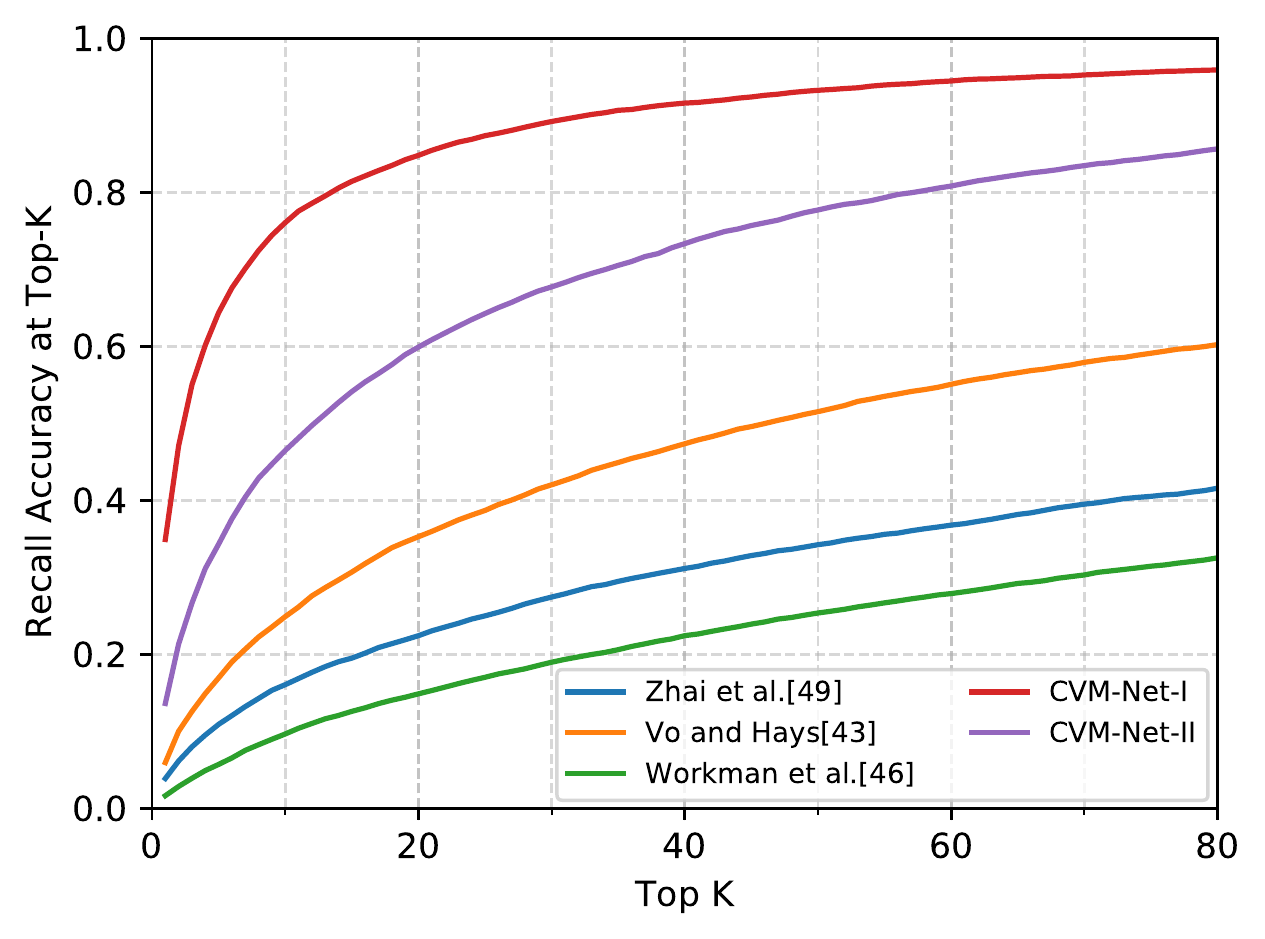}
		\caption {Comparison of our CVM-Nets and other existing approaches~(\cite{Zhai2017}; \cite{Vo2016}; \cite{Workman2015b}): All models are trained on CVUSA~\cite{Zhai2017}.}
		\label {fig:top_accuracy_curve}
	\end{figure}

\begin{figure*}
		\centering
		\includegraphics[width=0.97\linewidth]{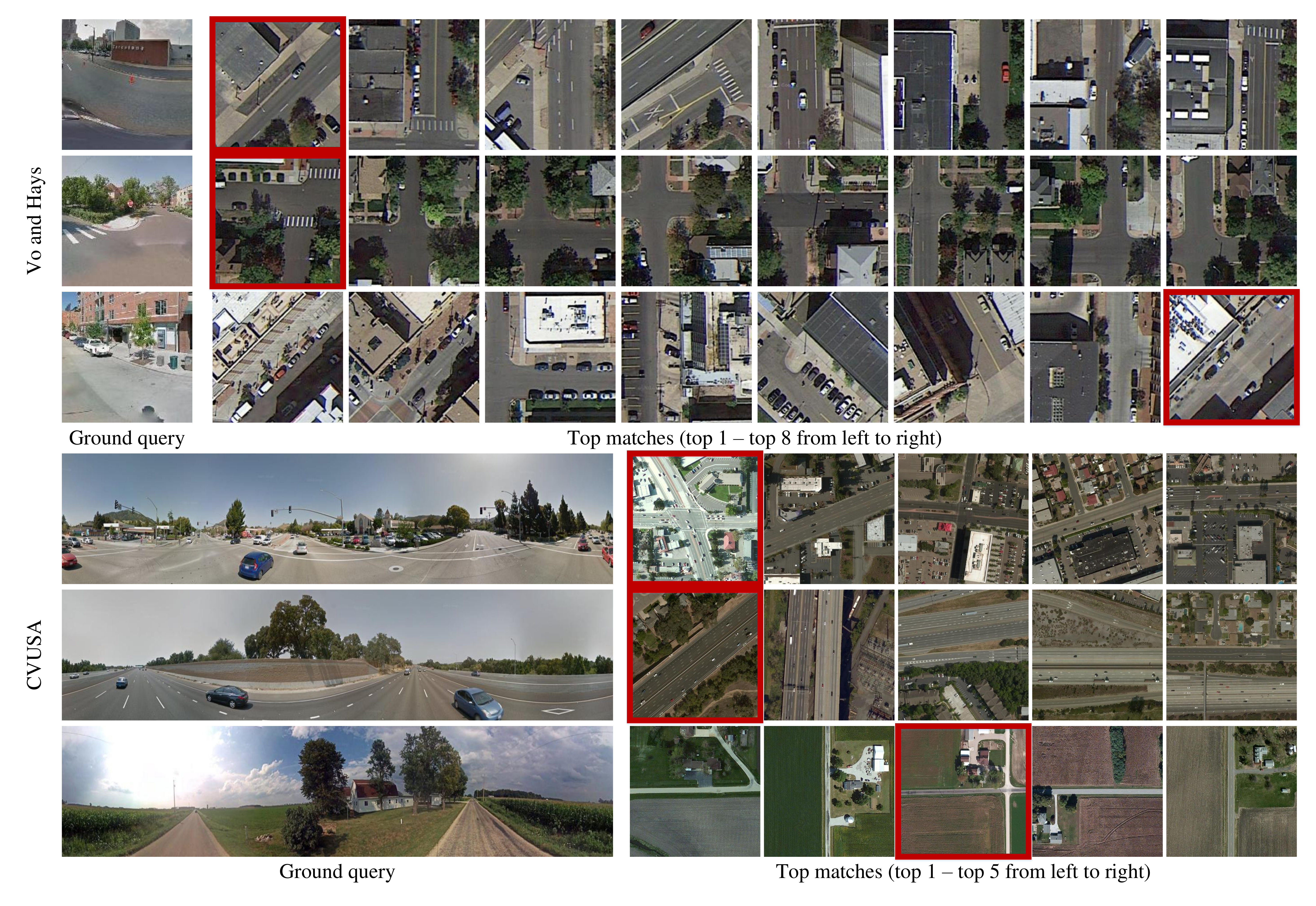}
		\caption {Image retrieval examples on Vo and Hays dataset~\cite{Vo2016} and CVUSA dataset~\cite{Zhai2017}. The satellite image bordered by red square is the groundtruth.}
		\label {fig:visual_retrieval}
	\end{figure*}

We show the recall accuracy from top 1 to top 80 (top~0.9\%) of our CVM-Nets with all the other approaches on CVUSA dataset~\cite{Zhai2017} in Figure~\ref{fig:top_accuracy_curve}. It illustrates that our proposed networks outperform all the other approaches. In Figure~\ref{fig:visual_retrieval}, we show some retrieval examples on two benchmark datasets~\cite{Vo2016} and \cite{Zhai2017}.

\paragraph{Local feature extraction architectures}
We evaluate our CVM-Net-I with different convolutional neural network for local feature extractions. Four commonly used neural network are compared: VGG~\cite{VGG}, ResNet~\cite{He2016ResNet}, Den-
\newline
senet~\cite{Huang2017DenseNet}, Xception~\cite{Chollet2017Xception}. Specifically, the convolutional parts of VGG-16, ResNet-50, DenseNet-121 ($k=32$) and Xception are used to extract local features of images. A 1 $\times$ 1 convolutional layer is added at the top to reduce the dimension of local feature vector to 512. All parameters are initialized with a pre-trained model on ImageNet~\cite{Deng2009}. The comparison results on the CVUSA dataset~\cite{Zhai2017} are shown in Table~\ref{tab:ablation_cnn_cvusa}. As can be seen from the table, the differences across different convolutional architectures on the top 1\% recall accuracy are marginal. It is interesting to note that VGG outperforms other architectures although they were shown to perform better in the classification tasks~\cite{He2016ResNet,Huang2017DenseNet,Chollet2017Xception}.

\begin{table}
	\centering
	\caption{Performance of different convolutional architectures on the CVUSA dataset \cite{Zhai2017}. The network is CVM-Net-I.}
	\begin{tabular}{|c|c|c|c|}
		\hline
		VGG    & ResNet  & DenseNet & Xception \\
		\hline
		96.3\% & 88.0\%  & 89.0\%   & 93.2\%   \\
		\hline
	\end{tabular}
	\label{tab:ablation_cnn_cvusa}
\end{table}

\begin{figure}
		\centering
		\includegraphics[width=0.8\linewidth]{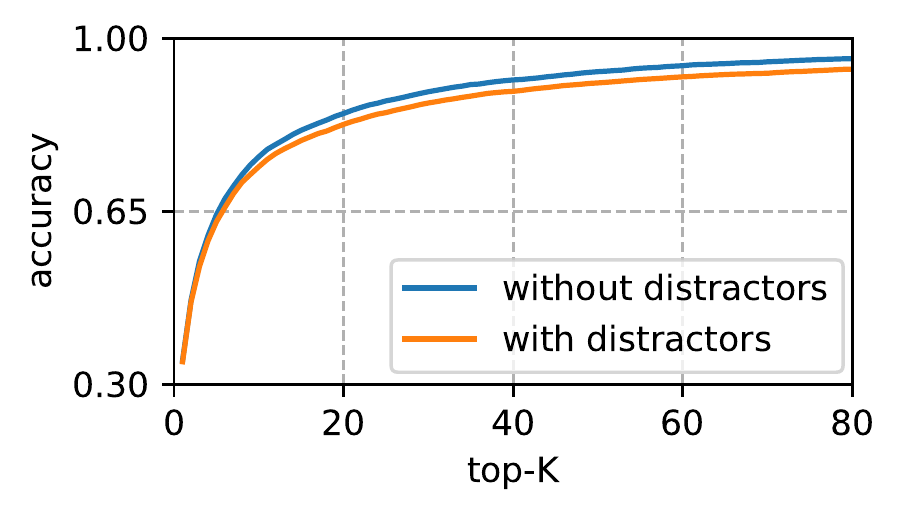}
		\caption {Top-K recall accuracy on the evaluation dataset with and without distractor images.The model is trained on CVM-Net-I on CVUSA dataset~\cite{Zhai2017}.}
		\label {fig:accuracy_expand_curve}
	\end{figure}

\paragraph{Adding distractor images}
We add 15,643 distractor satellite images in Singapore to our original test database which has 8,884 satellite images in USA. Figure~\ref{fig:accuracy_expand_curve} shows the top-K recall accuracy curve. The result is from the model trained on CVM-Net-I on the CVUSA~\cite{Zhai2017} dataset. There is only a marginal difference between the results with and without distractor images. This demonstrates the robustness of our proposed networks. 

\subsection{Discussions of CVM-Net}

\begin{table}
	\centering
	\caption{Performance of different architectures and losses on the CVUSA dataset \cite{Zhai2017}: AlexNet~\cite{AlexNet} and VGG16~\cite{VGG} are used as the local feature extraction network.}
	\begin{tabular}{l|ccc}
		\hline
		                       & Triplet & Quadruplet \\
		\hline \hline
		CVM-Net-I (AlexNet)    & 65.4\%  & 73.7\%     \\
		CVM-Net-I (VGG16)      & 96.3\%  & 89.9\%     \\
		CVM-Net-II (AlexNet)   & 63.0\%  & 83.9\%     \\
		CVM-Net-II (VGG16)     & 87.2\%  & 88.7\%     \\
		\hline
	\end{tabular}
	\label{tab:ablation_cvusa}
\end{table}

\paragraph{Local feature extraction}
In Table~\ref{tab:ablation_cnn_cvusa} and \ref{tab:ablation_cvusa}, we compare several variations on our proposed architecture. The deeper CNNs, i.e. VGG, ResNet, DenseNet and Xception significantly outperforms the shallower CNN, i.e. AlexNet. This result is not surprising because a deeper network is able to extract richer local features. However, an overly deep network does not necessarily generate better result. We observe a drop in the performances 
of the deeper networks - ResNet and DenseNet compared to the relatively shallower networks - VGG and Xception. 
This result suggests that a very deep convolutional network is not suitable for local feature extraction in the cross-view matching task despite its strong performances in the classification tasks. We reckon that this is because very deep networks extract high level features which is good for classification tasks, but might not be necessarily beneficial to our cross-view matching task due to the drastic change in viewpoint, where there is no similarity between the high level features across the different views.

\paragraph{CVM-Net-I vs CVM-Net-II}
It can be seen from Table~\ref{tab:ablation_cvusa} that CVM-Net-I outperforms CVM-Net-II on both the VGG16 and AlexNet implementations for local features extraction, and on both the triplet and quadruplet losses. This further reinforces our claim in the previous paragraph that shared weights implemented on CVM-Net-II is not necessarily good for our cross-view image-based retrieval task. 
We conjecture that CVM-Net-I outperforms CVM-Net-II because the aligned NetVLAD layers (i.e. two NetVLAD layers without weight sharing) have a higher capacity, i.e. more flexibility in having more weight parameters, in learning the features for cross-view matching.
In contrast, CVM-Net-II uses one shared fully connected layer on the input images that has limited capacity to transform local features from different domains into a common domain.
The comparison result from our experiment suggests that explicit use of the aligned NetVLADs is better than the naive use of fully connected layers on the cross-view matching task. 
Nonetheless, we propose both CVM-Net-I and CVM-Net-II in this paper. This is because we only conduct experiments on the cross-view image matching task, and we do not rule out the possibility that CVM-Net-II may outperform CVM-Net-I on other cross-domain matching tasks.  

\paragraph{Rotation and scale invariant}
Our proposed network can achieve rotation and scale invariant to some extent due to two reasons. First, the NetVLAD layer aggregates local features to a global descriptor regardless of the order in the local features. Hence, the rotation of the local feature maps from the rotated input image does not influence the global features. 
Second, we do training data augmentation. More specifically, we randomly rotate, crop and resize satellite images to make the network more robust on the change in rotation and scale.

\paragraph{Ranking loss}
The triplet loss has been widely used in image retrieval for a long time, while the quadruplet loss~\cite{Chen2017} was introduced recently to further improve the triplet loss. 
We train our CVM-Nets implemented with AlexNet and VGG16 for local feature extraction on both the triplet and quadruplet losses for comparison. As can be seen from the results in Table~\ref{tab:ablation_cvusa}, quadruplet loss outperforms triplet loss significantly on both our CVM-Nets with AlexNet. However, only minor differences in performances of the triplet and quadruplet losses can be observed for our CVM-Nets with VGG16. These results suggest that quadruplet loss has a much larger impact on shallower networks, i.e. AlexNet for feature extraction. 
We also train our CVM-Net-I and II on CVUSA dataset~\cite{Zhai2017} on the contrastive loss that was used in many earlier works. The top 1\% recall accuracy is 87.8\% and 79.8\% respectively. It is not as good as the results from the triplet loss or the quadruplet loss as shown in Table~\ref{tab:ablation_cvusa}. 

\begin{figure}
		\centering
		\includegraphics[width=0.8\linewidth]{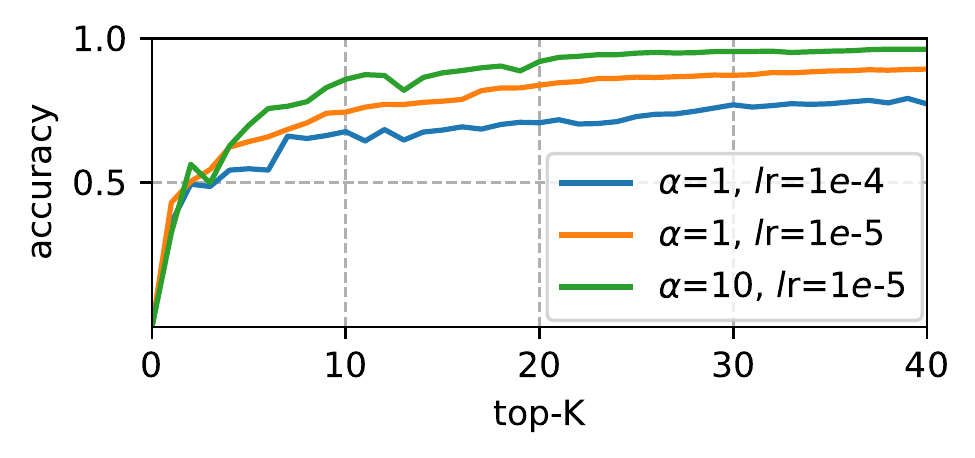}
		\caption {Performance of our weighted soft-margin triplet loss with different parameters. \textit{lr} is short for learning rate. It takes about 1 hour to train each epoch.}
		\label {fig:loss_compare}
	\end{figure}

\paragraph{Weighted soft-margin}
We also compare the performance of our CVM-Nets on different $\alpha$ values in our weighted soft-margin triplet loss $\mathcal{L}_{weighted}$ in Equation~\ref{eqn:our_loss}. Specifically, we conduct experiments on $\alpha=10$ with learning rate $10^{-5}$, $\alpha=1$ (soft-margin triplet loss) with learning rate $10^{-5}$. In addition, we also tested on $\alpha=1$ with learning rate $10^{-4}$ to compare the convergence speed with our weighted loss.
The accuracies from the respective parameters with respect to the number of epochs are illustrated in Figure~\ref{fig:loss_compare}. As can be seen, our loss function makes the network converges to higher accuracies in a shorter amount of time.  We choose $\alpha=10$ in our experiments since the larger value of $\alpha$ does not make much different.

\subsection{Image-Based Geo-Localization}

We choose the CVM-Net-I with weighted soft-margin triplet loss for the image-based geo-localization experiment. This is because experiment results from the previous section show that it gives the best performance for the ground-to-satellite image retrieval task.

\begin{figure*}
		\centering
		\includegraphics[width=0.97\linewidth]{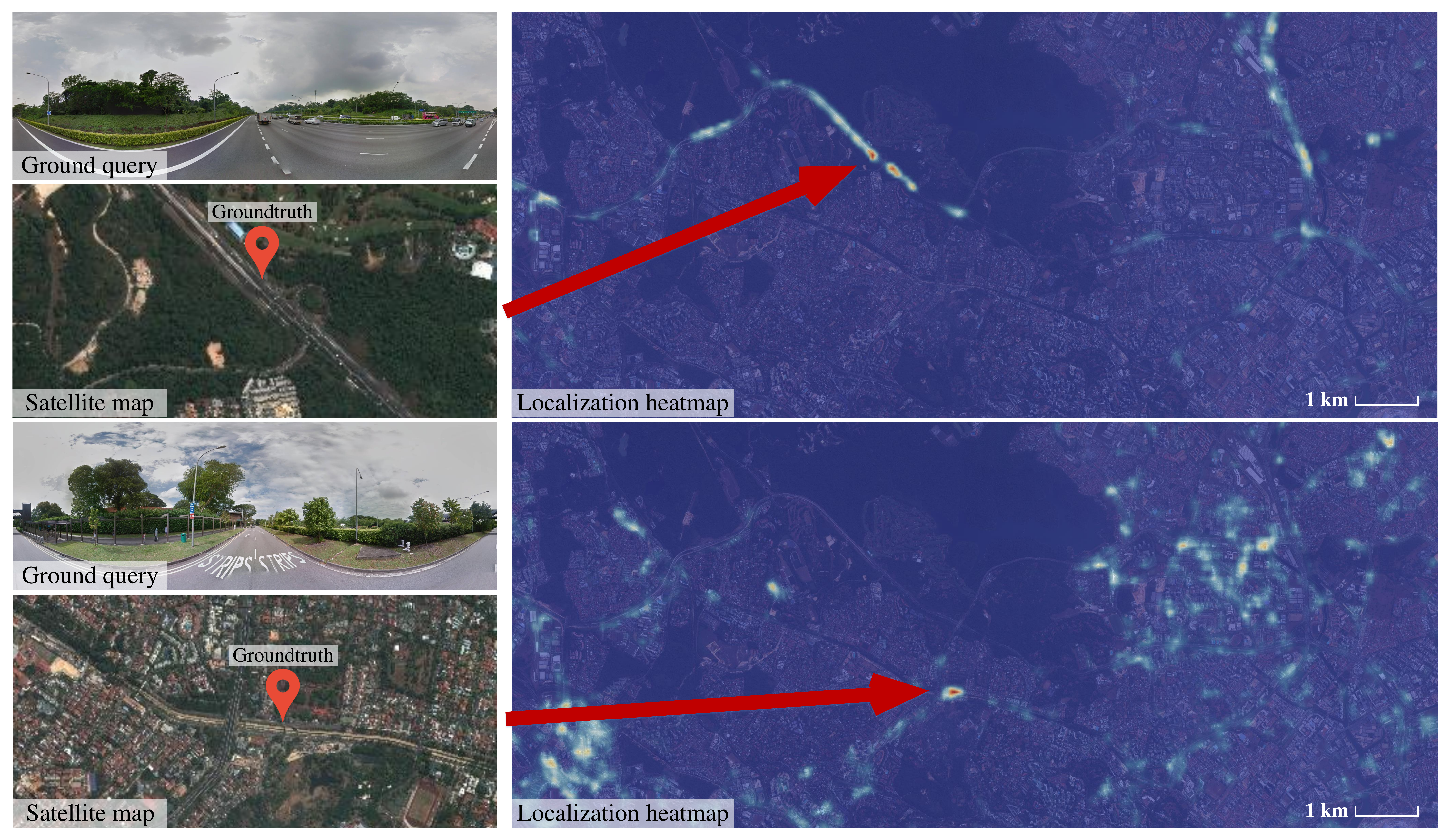}
		\caption {Large-scale geo-localization examples on our dataset.}
		\label {fig:localization}
	\end{figure*}

\paragraph{Without particle filter}
We perform image-based geo-localization with respect to a geo-referenced satellite map with our cross-view image retrieval CVM-Net. Our geo-referenced satellite map covers a region of $10 \times 5$ km of the South-East Asian country - Singapore. We collect the ground panoramic images of Singapore from Google Street-view. We choose to test our CVM-Net on Singapore to show that our CVM-Net trained on the North American based CVUSA datasets generalize well on a drastically different area. 
We tessellate the satellite map into grids at 5m intervals. Each image patch is $512 \times 512$ pixels and the latitude and longitude coordinates of the pixel center give the location of the image patch. 
We use our CVM-Net-I trained on the CVUSA dataset to extract global descriptors from our Singapore dataset. We visualize the heatmap of the similarity scores on the reference satellite map of two examples in Figure~\ref{fig:localization}. 
We apply the exponential function to improve the contrast of the similarity scores. It can be seen that our CVM-Net-I is able to recover the ground truth locations for both examples in Figure~\ref{fig:localization}.  
It is interesting to see that our street-view based query image generally return higher similarity scores on areas that correspond to the roads on the satellite map. 

\begin{figure}
		\centering
		\includegraphics[width=0.8\linewidth]{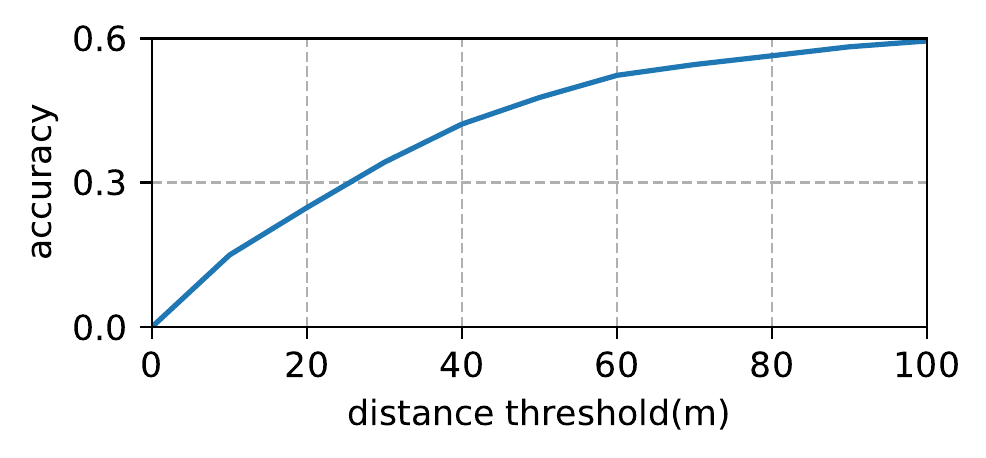}
		\caption {The retrieval accuracy on distance error threshold without particle filtering.}
		\label {fig:metric_accuracy_curve}
	\end{figure}

We conduct a metric evaluation on geo-localization. A query is regarded as correctly localized if the distance to the ground truth location is less than the threshold. We show the recall accuracy with respect to the distance threshold in Figure~\ref{fig:metric_accuracy_curve}. The accuracy on a 100m threshold is 67.1\%. The average localization error is 676.7m. As can be seen from the metric evaluation result, there is a large room for improvement in the ground-to-aerial geo-localization task despite our state-of-art retrieval performance. 
The localization accuracy of CVM-Net is not enough for the real-world applications. Our proposed Markov localization framework reduces the localization error and is evaluated on a real-world application.

\paragraph{With particle filter}
We perform the real-world experiment in two areas of Singapore - One North and South Buona Vista. We collect a small amount of data from our vehicle and use them to fine-tune the network trained on the CVUSA~\cite{Zhai2017} dataset. To accelerate the localization on the vehicle, the satellite map is discretized into a database of images. The descriptors of all images are pre-computed through our CVM-Net-I and stored offline. During the experiment, only ground view images need to be fed into the network. The initial pose of the vehicle is given from the GNSS/INS system.

\begin{table}
	\centering
	\caption{Average localization accuracy}
	\begin{tabular}{l|cc}
		\hline
		                   & Position (m) & Heading (degree) \\
		\hline \hline
		One North          & 16.39        & 0.25             \\
		South Bouna Vista  & 20.33        & 0.56             \\
		\hline
	\end{tabular}
	\label{tab:loc_avg_accuracy}
\end{table}

\begin{figure*}
		\centering
		\includegraphics[width=\linewidth]{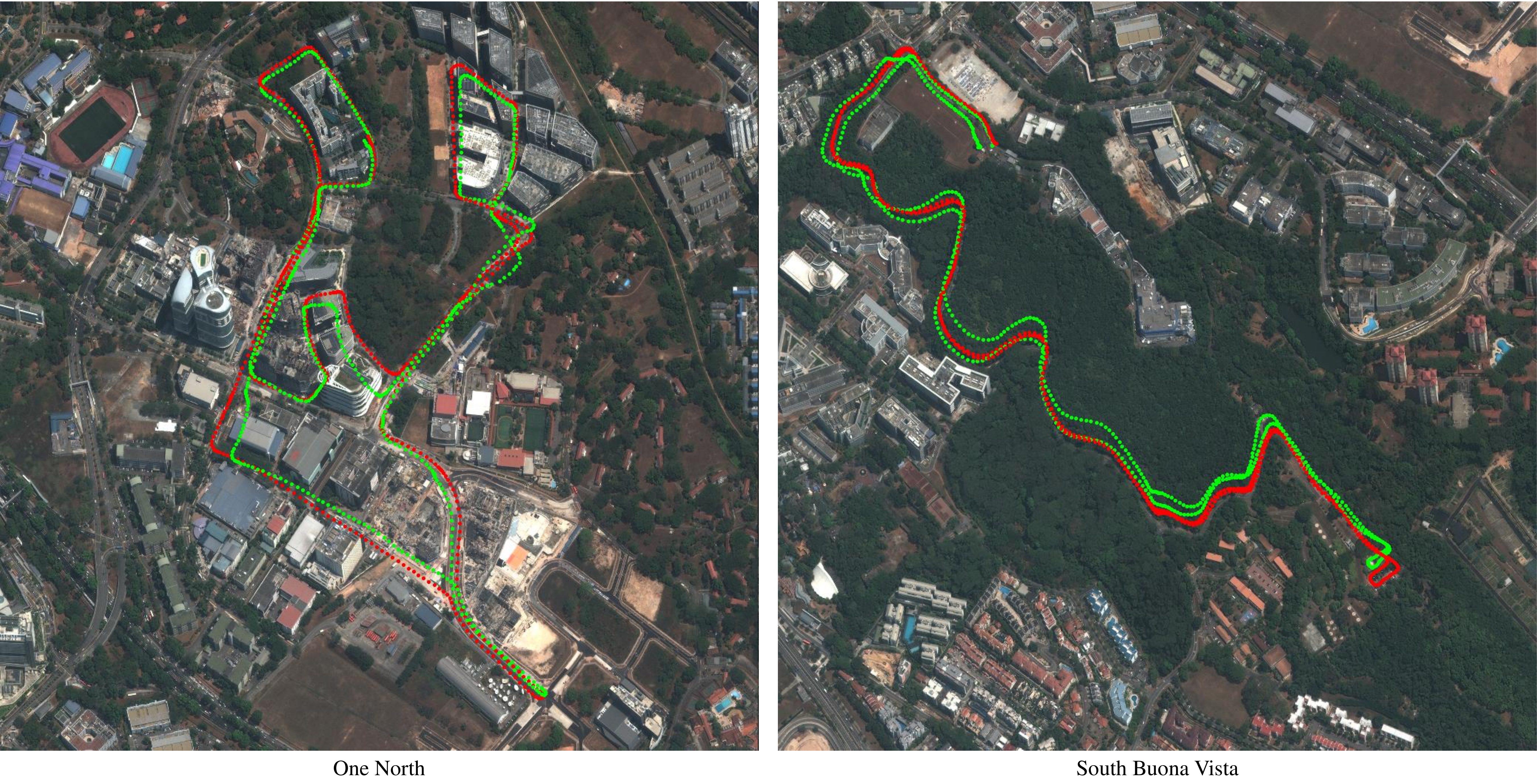}
		\caption {The localization results on One North and South Buona Vista. The red dots are ground-truth location and the green dots are the location estimated by our proposed framework. One North is an urban environment while South Buona Vista is a rural environment.}
		\label {fig:pf_result}
	\end{figure*}

Figure~\ref{fig:pf_result} shows the results of our image-based cross-view geo-localization framework executed live on the vehicle. The average error is shown in Table~\ref{tab:loc_avg_accuracy}. The position error is the Euclidean distance between the estimated position $[x_{est}, y_{est}]$ and the ground-truth position $[x_{gt}, y_{gt}]$:
	\begin{equation}
	\label{eqn:error_position}
		error_{pos} = \sqrt{(x_{est} - g_{gt})^2 + (y_{est} - y_{gt})^2}.
	\end{equation}
The heading error is the difference between the estimated heading and the ground-truth heading. We use the atan2 function to compute the angle difference to prevent the wrap-around problem:
	\begin{equation}
		error_{\theta} = atan2(v_{est}, v_{gt}).
	\end{equation}
\noindent $v_{est}$ is the heading unit vector of the estimated heading $\theta_{est}$ and $v_{gt}$ is the heading unit vector of the ground-truth heading $\theta_{gt}$. The total length of trajectory in One North is about 5km and the length of trajectory in South Bouna Vista is about 3km. From the results, it can be seen that our proposed framework can localize the vehicle along a long path within a small error in both the urban area and the rural area. The localization frequency is around 0.5Hz to 1Hz.

\section{Conclusion}
\label{sec:conclusion}

In this paper, we introduce two cross-view matching networks - CVM-Net-I and CVM-Net-II, which are able to match ground view images with satellite images in order to achieve cross-view image localization. We introduce the weighted soft-margin ranking loss and show that it notably accelerates training speed and improves the performance of our networks. Furthermore, we propose a Markov Localization framework that fuses the satellite localization and visual odometry to localize the vehicle. We demonstrate that our proposed CVM-Nets significantly outperforms state-of-the-art approaches with experiments on large datasets. We show that our proposed framework can continuously localize the vehicle within a small error.


\bibliographystyle{spmpsci}      

\end{document}